\documentclass{article}
\usepackage{ijcai25}

\usepackage{times}
\usepackage{soul}
\usepackage{url}
\usepackage[hidelinks]{hyperref}
\usepackage[utf8]{inputenc}
\usepackage[small]{caption}
\usepackage{graphicx}
\usepackage{amsmath}
\usepackage{amsthm}
\usepackage{booktabs}
\usepackage{algorithm}
\usepackage{algorithmic}
\usepackage[switch]{lineno}
\usepackage{amssymb}
\usepackage{comment}
\usepackage{multirow}
\usepackage {subcaption}
\usepackage{color}

\newcommand{\ie}{\textit{i}.\textit{e}.}
\newcommand{\eg}{\textit{e}.\textit{g}.}

\title{Leveraging MLLM Embeddings and Attribute Smoothing for \\ Compositional Zero-Shot Learning}

\author{
Xudong Yan$^1$
\and
Songhe Feng$^1$\thanks{Corresponding authors}\and
Yang Zhang$^1$\and
Jian Yang$^2$\and
Yueguan Lin$^2$\and
Haojun Fei$^{2*}$
\\
\affiliations
$^1$School of Computer Science and Technology, Beijing Jiaotong University\\
$^2$Qifu Technology\\
\emails
\{xud\_yan,  shfeng,  chefzhang\}@bjtu.edu.cn,
\{yangjian1,  linyueguan,  feihaojun\}-jk@360shuke.com
}
\begin{document}

\maketitle
\begin{abstract}
Compositional zero-shot learning (CZSL) aims to recognize novel compositions of attributes and objects learned from seen compositions. Previous works disentangle attributes and objects by extracting shared and exclusive parts between the image pair sharing the same attribute (object), as well as aligning them with pretrained word embeddings to improve unseen attribute-object recognition. Despite the significant achievements of existing efforts, they are hampered by three limitations: (1) The efficacy of disentanglement is compromised due to the influence of the background and the intricate entanglement of attributes with objects in the same parts. (2) Existing word embeddings fail to capture complex multimodal semantic information. (3) Overconfidence exhibited by existing models in seen compositions hinders their generalization to novel compositions. Being aware of these, we propose a novel framework named multimodal large language model (MLLM) embeddings and attribute smoothing guided disentanglement for CZSL. First, we leverage feature adaptive aggregation modules to mitigate the impact of background, and utilize learnable condition masks to capture multi-granularity features for disentanglement. Moreover, the last hidden states of MLLM are employed as word embeddings for their superior representation capabilities. Furthermore, we propose attribute smoothing with auxiliary attributes generated by the large language model (LLM) for seen compositions to address the overconfidence challenge. Extensive experiments demonstrate that our method achieves state-of-the-art performance on three challenging datasets. The supplementary material and source code will be available at \url{https://github.com/xud-yan/Trident}.
\end{abstract}
\section{Introduction}
As for the study of compositional generalization ability inherent in human beings, compositional zero-shot learning (CZSL)~\cite{2017_Misra_CVPR_first,2018_Nagarajan_ECCV_AttrAsOp,2019_Purushwalkam_ICCV_TMN} 
is proposed to enable machines to recognize unseen compositions by leveraging knowledge of attributes and objects (\ie, primitives) learned from seen compositions. Specifically, in the training phase, the models are provided with images of seen compositions (\eg, \texttt{ripe orange} and \texttt{peeled apple}). During the testing phase, given an image that depicts a novel composition (\eg, \texttt{peeled orange}), models are assigned to correctly recognize it~\cite{2022_Zhang_ECCV_INV}.

Prior works~\cite{2017_Misra_CVPR_first,2019_Nan_AAAI,2021_Naeem_CVPR_CGE} focus on mapping the visual features and the word embeddings of compositions into a joint space. 
These methods have poor generalization capabilities on unseen compositions due to the recombination of primitives. 
Therefore, recent studies~\cite{2022_Saini_CVPR_OADis,2023_Wang_CVPR_CANET,2024_Li_TCDS,2024_IJCAI_CCZSL} consider visual disentanglement. 
Among them, some prominent works~\cite{2023_Hao_CVPR_ADE} deploy a triplet of images to disentangle visual features: a given image and two supplementary images, each sharing either the same attribute or object as the given image. The triplet of images is treated as two image pairs for subsequent analysis. 
These approaches aim to disentangle attribute and object by extracting the shared and exclusive features of the image pair, as well as aligning them with word embeddings (\eg, GloVe~\cite{2014_Pen_EMNLP_glove}), as shown in Figure~\ref{fig:vice-pipeline}. 
Although these pioneering research studies have made great progress, they exhibit three limitations: 

\begin{figure*}[!t]
    \centering
    \includegraphics[width=0.95\linewidth]{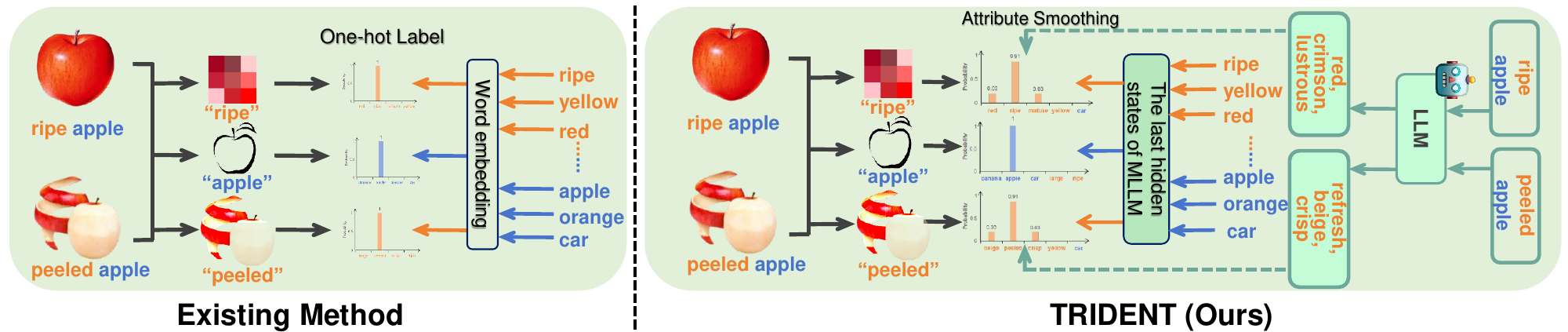}
    \vspace{-0.5em}
    \caption{A general comparison between the existing method and our proposed \textbf{TRIDENT}. Note that, we only present the representation learning of an image pair sharing the object for brevity.}
    \label{fig:vice-pipeline}
\end{figure*}

\textbf{L1:} Disentanglement is impeded due to the influence of the background and the intricate entanglement between attributes and objects in the same parts of images. 
On the one hand, models tend to extract the background features unique to one image in the pair as the disentangled exclusive features. 
On the other hand, some existing methods~\cite{2021_Ruis_NIPS_ProtoProp,2022_Saini_CVPR_OADis} compute the similarity of the image pair for disentanglement at the spatial level. However, this paradigm is limited by the frequent entanglement of attributes and objects within the same image regions.

\textbf{L2:} Existing word embeddings lack the depth needed to capture complex multimodal semantic information. 
To begin with, word embeddings (\eg, 
GloVe~\cite{2014_Pen_EMNLP_glove}) are grounded in word frequency and contextual co-occurrence, overlooking the high-level semantic nuances~\cite{2021_sarzynska_detecting}. 
Moreover, the process of aligning visual features with word embeddings can be viewed as a form of cross-modal matching; however, these word embeddings are trained only in a single text modality, failing to capture multimodal information between images and texts.

\textbf{L3:} Existing methods display excessive confidence in seen compositions, impairing their ability to generalize towards novel compositions. 
Specifically, due to the one-hot label used during training, these approaches are limited by learning only one disentangled attribute and object, neglecting the fact that objects naturally exhibit multiple attributes~\cite{2024_xu_arxiv_mac}. 
Consequently, models exhibit overconfidence in the disentangled ground-truth attribute, treating other attributes that can describe the object as negative ones, which results in the diminished generalization to unseen compositions.

Being aware of these, we propose a novel framework named
multimodal large language model (MLLM) embeddings and at\textbf{TR}ibute smooth\textbf{I}ng gui\textbf{DE}d dise\textbf{NT}anglement (\textbf{TRIDENT}), which consists of three major modules: visual feature extraction, attribute-object disentanglement, and feature alignment. 
The first module leverages feature adaptive aggregation (FAA) modules to mitigate the impact of background noise, and exploits learnable condition masks for multi-granularity feature learning 
at the dimensional level 
to improve subsequent disentanglement. 
The second module aims at leveraging shared and exclusive weights of image pairs to disentangle attributes and objects under the paradigm that apart from the shared features of the image pair, each image has its own exclusive features. 
The third module is intended to align the visual features of compositions and disentangled primitives with the last hidden states of MLLM (\ie, MLLM embeddings). 
This is inspired by prior works~\cite{2020_wang_ASLP_SBERT,2022_Mu_arxiv_sgpt,2024_muennighoff_GritLM}, which reveal that the last hidden states of (M)LLM exhibit powerful representation capabilities in embedding tasks (\eg, retrieval and classification). 
Moreover, to tackle the issue that the overconfidence of the models regarding the ground-truth attribute hinders them from generalizing to unseen compositions, we exploit the large language model (LLM) to generate auxiliary attributes for compositions and perform label smoothing for attributes (\ie, attribute smoothing).

In summary, the contributions of our work are three-fold:

1. We propose novel feature adaptive aggregation modules to reduce the impact of background, and utilize learnable condition masks to capture multi-granularity features at the dimensional level for disentanglement in CZSL.

2. We employ both LLM and MLLM to guide attribute-object disentanglement by generating auxiliary attributes and representing primitive words for CZSL, respectively.

3. Extensive experiments conducted on three challenging datasets (MIT-States~\cite{2015_Isola_CVPR_mit}, C-GQA~\cite{2021_Naeem_CVPR_CGE}, and VAW-CZSL~\cite{2022_Saini_CVPR_OADis}) show that \textbf{TRIDENT} has achieved state-of-the-art performance.
\section{Related Work}
\begin{figure*}[t]
    \centering
    \includegraphics[width=\linewidth]{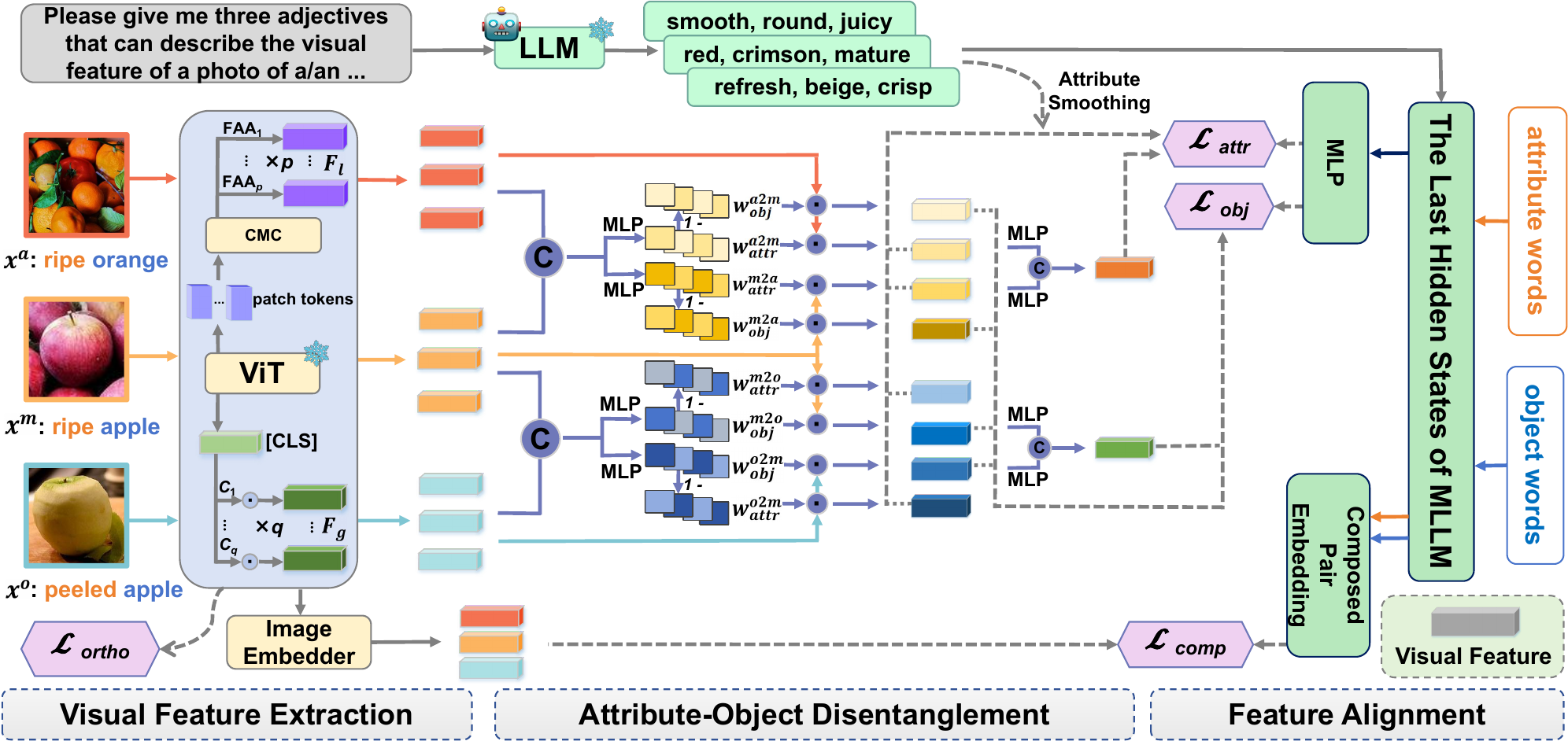}
    \caption{The overall architecture of our proposed \textbf{TRIDENT}. The model consists of three major modules: (a) visual feature extraction, (b) attribute-object disentanglement, and (c) feature alignment.}
    \label{fig:main-pipeline}
\end{figure*}
\textbf{Compositional zero-shot learning (CZSL).}
Prior works in CZSL can be broadly divided into two main streams. 
One main stream is learning visual representations and textual labels of compositions in a joint space. 
SymNet~\cite{2020_Li_CVPR_SymNet} aims to learn the symmetry property in compositions. 
Co-CGE~\cite{2022_Mancini_PAMI_CoCGE} leverages a graph convolutional neural network to learn composition representations. 
The other main stream aims at disentangling visual representations of primitives to reduce composition learning to primitive learning. 
SCEN~\cite{2022_Li_CVPR_SCEN} leverages contrastive loss to excavate discriminative prototypes of primitives. 
CANet~\cite{2023_Wang_CVPR_CANET} learns the conditional attribute conditioned on the recognized object and the input image. 
More recent works~\cite{2023_Saini_ICLR_CSP,2023_Lu_CVPR_DFSP,2024_Huang_CVPR_Troika} leverage the encyclopedic knowledge of pre-trained vision-language models (VLM) like CLIP~\cite{2021_Radford_ICML_CLIP} to encode and align images and texts. 

\textbf{Large language model (LLM).} 
LLMs have realized significant advancements thanks to the scaling up of training data and the increase in the number of parameters. 
Early models, such as GPT-2~\cite{2019_Radford_Arxiv_GPT2}, initially exhibit strong capabilities in understanding and generating human-like language. 
Subsequently, GPT-3~\cite{2020_Brown_NIPS_GPT3} and LLaMA~\cite{2023_touvron_arxiv_llama} demonstrate great breakthroughs across numerous language benchmarks.

Expanding on LLMs, multimodal large language models (MLLM) incorporate a visual encoder for vision-language tasks.
Flamingo~\cite{2022_NIPS_Flamingo} integrates Vision Transformer (ViT)~\cite{2021_Dosovitskiy_Vit} and LLM by gated cross-attention. 
LLaVA~\cite{2024_liu_NIPS_llava} and LLaVA v1.5~\cite{2024_liu_CVPR_llava1.5} introduce visual instruction tuning to enhance the instruction following capability. The visual understanding part of LLaVA v1.5 consists of a ViT and an MLP cross-modal connector. 
We choose LLaVA v1.5 as our foundational MLLM for its state-of-the-art performance.

Recently, exploring the powerful embedding capabilities of (M)LLM to handle representation tasks (\eg, retrieval) has emerged as a prominent research domain.
SGPT~\cite{2022_Mu_arxiv_sgpt} exploits the last hidden states of LLM for the input token sequence or a special learnable token to derive representational embeddings. Subsequently, GritLM~\cite{2024_muennighoff_GritLM} applies mean pooling over the last hidden states of LLM to produce the textual embeddings. 
\section{Approach}
\subsection{Task Formulation}
Compositional zero-shot learning (CZSL) aims at learning a model that can recognize unseen compositions of attributes and objects that are learned from seen compositions. Given an attribute set $\mathcal{A}$ and an object set $\mathcal{O}$, the attributes and objects are composed to form a composition set $\mathcal{C} = \mathcal{A} \times \mathcal{O}$. The composition set $\mathcal{C}$ is divided into two disjoint sets: the seen composition set $\mathcal{C}_s$ and the unseen composition set $\mathcal{C}_u$, \ie, $\mathcal{C}_s \cap \mathcal{C}_u = \varnothing$. The model learns from a seen training set $\mathcal{D}_{tr} = \{(x_s, c_s)\}$, where $x_s$ is an image of the seen composition label $c_s \in C_s$. Following the Generalized CZSL~\cite{2019_Purushwalkam_ICCV_TMN}, the model is evaluated on a predefined test set $\mathcal{D}_{te} = \{(x_{te}, c_{te})\}$, where $x_{te}$ is a test image from the predefined composition subset $\mathcal{C}_{te}$ of $\mathcal{C}$, \ie, $\mathcal{C}_{te} \subseteq \mathcal{C}$, and $c_{te} \in \mathcal{C}_{te}$ is the label of $x_{te}$. 

\subsection{\textbf{TRIDENT}}
As the major novelty, we propose a novel framework named MLLM embeddings and attribute smoothing guided disentanglement (\textbf{TRIDENT}), as shown in Figure~\ref{fig:main-pipeline}. It consists of three modules: (1) visual feature extraction, (2) attribute-object disentanglement, and (3) feature alignment. 

\subsubsection{Visual Feature Extraction}
\label{sec:visual-feature-extraction}
As shown in Figure~\ref{fig:main-pipeline}, we denote a given image with the attribute-object composition label (e.g. \texttt{ripe apple}) as the main image $x^m$, and randomly sample an image with the same attribute $x^{a}$ (\ie, \texttt{ripe orange}) and an image sharing the same object $x^{o}$ (\ie, \texttt{peeled apple}) to form a triplet image set. For convenience of expression, we simply use $x^{img}$ (where $img \in \{m, a, o\}$) to collectively denote the images as they are processed using the same module.

\textbf{Visual backbone.} As mentioned before, since LLaVA v1.5 is used as our fundamental MLLM, we directly leverage its visual encoder (\ie, ViT) and cross-modal connector (CMC) to extract visual features. Specifically, the image $x^{img}$ is partitioned into $n$ patch tokens, which are subsequently put into ViT along with the \texttt{[CLS]} token. 
Afterward, the output of patch tokens before the last layer of ViT is fed into the CMC module, as implemented in LLaVA v1.5. 
To align the dimensions of the patch tokens output by CMC and the \texttt{[CLS]} token produced by ViT, the patch tokens output by CMC is input into a linear layer. 
Consequently, we obtain one feature vector of \texttt{[CLS]} token $f^{img}_{cls} \in \mathbb{R}^{d}$ and a patch feature matrix of $n$ patch tokens $F^{img}_{patch} \in \mathbb{R}^{n \times d}$.

\textbf{Local features extraction.} Intuitively, the composition (\eg, \texttt{ripe apple}) only occupies a few parts of the image. Since each patch token corresponds to one local region of the image, to filter out background noise and focus on related regions, we deploy $p$ feature adaptive aggregation (FAA) modules to derive $p$ relevant local features of $x^{img}$, where each FAA module is formulated as follows:
\begin{equation}
    \left\{ 
    \begin{array}{ll}
    v = agg \otimes F^{img}_{patch} \\[0.5em] agg = ReLU(Conv(F^{img}_{patch}))
    \end{array}
    \right.
\end{equation}
where $Conv(\cdot)$ represents the 1 $\times$ 1 convolution layer, $agg \in \mathbb{R}^{n}$ is the weight vector, and the $k$-th element of $agg$ is the weight for $k$-th patch feature. $\otimes$ represents matrix product, and $v \in \mathbb{R}^{d}$ is the local feature obtained by an FAA module. We vertically concatenate the local features produced by $p$ FAA modules to obtain the local feature matrix $F^{img}_l \in \mathbb{R}^{p \times d}$.

\textbf{Global features extraction.} Normally, the \texttt{[CLS]} token output by ViT is regarded as containing various global information of the image, which highly entangles both attribute and object together~\cite{2023_Hao_CVPR_ADE}.
To disperse multi-granularity global information into different representations at the dimensional level,
$q$ learnable condition masks are applied to $f^{img}_{cls}$ to obtain $q$ different global representations. Each global representation is computed as:
\begin{equation}
    u = f^{img}_{cls} \odot c
\end{equation}
where $u \in \mathbb{R}^{d}$ denotes the global representation, $c \in \mathbb{R}^{d}$ refers to the learnable condition mask and $\odot$ is the element-wise multiplication. We vertically concatenate $q$ global representations to derive the global feature matrix $F^{img}_g \in \mathbb{R}^{q \times d}$.

\textbf{Features concatenation.} Finally, $F^{img}_l$ and $F^{img}_g$ are vertically concatenated to form the visual features of $x^{img}$, \ie, $F^{img} = [F^{img}_l, F^{img}_g] \in \mathbb{R}^{h \times d}$ (where $h=p+q$), which is used for the following attribute-object disentanglement.

\textbf{Orthogonal regularization.} Different features should capture distinct and complementary aspects of the image. Therefore, we introduce the orthogonal regularization, i.e.:
\begin{small}
\begin{equation}
    \mathcal{L}_{ortho} = 
    \frac{1}{|img|\cdot|i|}
     \sum_{\substack{img \in \{m, a, o\} , i \in \{g, l\}}} (|| F^{img}_i{F^{img}_i}^T  - I_i ||_{Fro})
\end{equation}
\end{small}where $I_i$ denotes the identity matrix, and $||\cdot||_{Fro}$ refers to the Frobenius norm of the matrix.

\textbf{Image embedder.} 
Inspired by~\cite{2018_Nagarajan_ECCV_AttrAsOp},
for the input image $x^{img}$, we first use the average pooling $Avg(\cdot)$ on $F^{img}_g$ and $F^{img}_l$, respectively, and horizontally concatenate them by $Cat(\cdot, \cdot)$ to aggregate both global and local features of $x^{img}$. Then the concatenated feature passes through a linear layer $Lin_{comp}(\cdot)$ to derive the final visual feature $f^{img}_{comp}$ that represents the composition. This module is formulated as follows:
\begin{equation}
    f^{img}_{comp} = Lin_{comp}(Cat(Avg(F^{img}_g), Avg(F^{img}_l)))
\end{equation}

\subsubsection{Attribute-Object Disentanglement}
As mentioned before, one of the key challenges for CZSL is to disentangle attributes and objects from visual features. To overcome the challenge, we propose a novel weighted disentanglement strategy, as illustrated in Figure~\ref{fig:main-pipeline}. 
For brevity, the image pair $(x^m,x^a)$ from the triplet image set is taken as an example to elaborate on this strategy, while another image pair $(x^m, x^o)$ is processed in the same manner. 

\textbf{Weights computation.} The features of $x^m$ and $x^a$ (\ie, $F^m$ and $F^a$) are vertically concatenated and fed into two MLP modules to derive their respective weights of shared attribute features relative to each other. Subsequently, we utilize them to compute the weights of their own exclusive object features as follows:
\begin{equation}
    \left\{
    \begin{array}{ll}
        w_{attr}^{m2a} =  \sigma(MLP_{m2a}([F^m, F^a])) \\[0.5em]
        w_{obj}^{m2a} = \mathbf{1} - w_{attr}^{m2a} \\[0.5em]
        w_{attr}^{a2m} =  \sigma(MLP_{a2m}([F^m, F^a]))\\[0.5em]
        w_{obj}^{a2m} = \mathbf{1} - w_{attr}^{a2m}
    \end{array}
    \right.
\end{equation}
where $w_{attr}^{m2a}, w_{attr}^{a2m} \in \mathbb{R}^{h}$ represent the weights of the shared attribute features of $x^m$ relative to $x^a$ and $x^a$ relative to $x^m$, respectively. $w_{obj}^{m2a}$ and $w_{obj}^{a2m}$ denote the weights of exclusive object features corresponding to $x^m$ and $x^a$, respectively, 
which are derived by "$1 - shared\ weights$" paradigm as beyond the shared features of the image pair are the exclusive features of each image.
Taking $w_{attr}^{m2a}$ as an example, its $k$-th element refers to the shared attribute proportion of the $k$-th feature of $x^m$ relative to $x^a$.

\textbf{Disentangled features acquisition.} We multiply the elements of each weight vector by the corresponding features and then calculate the average. The following takes the process of obtaining the shared attribute features of image $x^{m}$ relative to $x^{a}$ as an example:
\begin{equation}
    f_{attr}^{m2a} = \frac{1}{h} \sum_{i=1}^{h} {w^{m2a}_{attr}}_i\ {F^m}_{i, :}
\end{equation}
where ${F^m}_{i, :}$ denotes the $i$-th row of $F^m$, \ie, the $i$-th feature of $x^m$. ${w^{m2a}_{attr}}_i$ refers to the $i$-th element of $w^{m2a}_{attr}$.

For the image pair of $x^m$ and $x^a$, four parts are obtained: the shared attribute features of $x^m$ relative to $x^a$ and $x^a$ relative to $x^m$, as well as two exclusive object features of the two images, respectively. 
These four features are marked as $f_{pri}^{e}$, where $e \in \{m2a, a2m\}$ and $pri \in \{attr, obj\}$.
Then the shared attribute feature of $x^a$ and $x^m$ without relativity is obtained by an MLP layer, which is less dependent on the different objects of the two images. The process is as follows:
\begin{equation}
    f_{attr}^{ma} = MLP_{ma}(Cat(f_{attr}^{m2a}, f_{attr}^{a2m}))
\end{equation}

Similarly, we disentangle the attribute and object for the image pair ($x^m$, $x^o$) and obtain the same visual features as ($x^m$, $x^a$): $f_{pri}^{e}$, where $e \in \{m2o, o2m\}$ and $pri \in \{obj, attr\}$, and the feature without relativity $f_{obj}^{mo}$.

\subsubsection{Feature Alignment} 
Inspired by GritLM~\cite{2024_muennighoff_GritLM} that leverages the last hidden states of LLMs as the representational embeddings, we consider the last hidden states of LLaVA v1.5~\cite{2024_liu_CVPR_llava1.5} as our MLLM embeddings for primitive words. 
Moreover, to tackle the problem that the ineffective overconfidence exhibited by the model in terms of the ground-truth attribute hinders it from generalizing to unseen compositions, we employ GPT-3.5~\cite{2023_openai_gpt35} to generate auxiliary attributes for compositions and perform label smoothing during attribute alignment. The auxiliary attributes and MLLM embeddings are obtained offline before training TRIDENT.

\textbf{Auxiliary attributes generation by LLM.} 
Since only textual attributes need to be generated, the LLM GPT-3.5~\cite{2023_openai_gpt35}, instead of an MLLM, is leveraged to generate $t$ auxiliary attributes for each composition. Specifically, the following prompt is input to LLM: '\textit{Please give me $t$ adjectives that can describe the visual feature of a photo of a/an ... well.}', where the attribute-object composition (\eg, \texttt{peeled apple}) is filled in '\textit{...}'. 
Subsequently, the generated auxiliary attribute words form a set $\mathcal{A}_a$. Therefore, the set of all words $\mathcal{Y}$ is obtained, including attributes, objects and auxiliary attributes:
\begin{equation}
    \mathcal{Y} = \mathcal{A} \cup \mathcal{O} \cup \mathcal{A}_a
\end{equation}

\textbf{MLLM embeddings acquisition.} 
Each word $y \in \mathcal{Y}$ is fed into LLaVA v1.5 to get the last hidden states, \ie, $LLaVA_{lhs}(\cdot)$. Specifically, $y$ is tokenized into multiple sub-words and passed through the MLLM; the final-layer output is averaged as the MLLM embedding of the word. 
Subsequently, it is passed through an MLP layer to obtain the embedding $E_{word}(\cdot)$ of the aligned dimension with visual features. 
And for a composed pair $c$ of attribute $a$ and object $o$, \ie, $c = (a, o)$, we horizontally concatenate the MLLM embeddings for $a$ and $o$ and feed them into a linear layer $Lin_{co}(\cdot)$ to get the composed pair embedding $E_{co}(\cdot)$. The process is formulated as follows:
\begin{equation}
    E_{word}(y) = MLP_{word}(LLaVA_{lhs}(y))
\end{equation}
\begin{equation}
    E_{co}(c) = Lin_{co}(Cat(LLaVA_{lhs}(a), (LLaVA_{lhs}(o)))
\end{equation}

\textbf{Word expanding.} Previous works compute cosine similarities of disentangled features and word embeddings and apply cross-entropy only within the respective domains of attributes or objects, which results in the disentangled attributes and objects still retaining the information of each other. To address the problem, we propose a novel word expanding strategy, which computes cosine similarities of visual features and the embeddings of all words, including attributes and objects, and treats all words except the ground-truth word as the negative labels in subsequent cross-entropy.

\textbf{Alignment by cross-entropy.} Similar to~\cite{2021_Mancini_CVPR_Compcos}, we use cross-entropy to align the visual features and word embeddings. Assume that $f$ is the visual feature and $E_{word}(wd)$ is the word embedding for the word $wd \in \mathcal{Y}$. The classifier logit from $f$ to $E_{word}(wd)$ is defined as follows:
\begin{equation}
    CE(f, wd)=\frac{e^{ \delta \cdot \cos(f, E_{word}(wd))}}{\sum_{wd^{'} \in \mathcal{Y}} e^{\delta \cdot \cos(f, E_{word}(wd^{'}))}}
\end{equation}
where $\delta$ is the temperature variable, and $cos(\cdot, \cdot)$ denotes the cosine similarity function. Thus cross-entropy with/without label smoothing can be uniformly formulated as follows:
\begin{equation}
   \begin{aligned}
    H(f, \mathcal{Y}) = \sum_{y \in \mathcal{Y}} -z \log(CE(f, y)) \\
    \text{with} \ \ z = 
    \begin{cases} 
        1 - \alpha, & \text{if } y \text{ is ground truth label} \\
        \alpha / t, & \text{if } y \text{ is auxiliary label} \\
        0, & \text{otherwise}
    \end{cases}
    \end{aligned}
\end{equation}
where $\alpha$ denotes the smoothing factor, $t$ refers to the number of auxiliary labels, and $z \in [0, 1]$ represents the target value of one-hot label or smoothing label. For cross-entropy without label smoothing, \ie, with one-hot label $H_{oh}$, $\alpha$ is set to $0$. The cross-entropy with label smoothing is denoted as $H_{ls}$. 

For the disentangled attribute features of one image relative to each other, since a single object exhibits multiple attributes, we exploit attribute smoothing with auxiliary attributes to undermine the overconfidence in the ground-truth attribute and learn more related attributes. For the shared attribute features without relativity, the one-hot label is used to learn a pure attribute concept that is less conditioned on objects. The loss for disentangled attributes is as follows:
\begin{small}
\begin{equation}
\begin{aligned}
\mathcal{L}_{attr} =
\frac{1}{|e|+1} ( 
\sum_{\scalebox{0.55}{$e \in \{m2a, a2m, m2o, o2m\}$}}{} H_{ls}(f_{attr}^{e}, \mathcal{Y}) +
H_{oh}(f_{attr}^{ma}, \mathcal{Y})
)
\end{aligned}
\end{equation}
\end{small}

Concerning the disentangled object features, we use cross-entropy with one-hot label to learn the object concept and the loss is as follows:
\begin{small}
\begin{equation}
\begin{aligned}
\mathcal{L}_{obj} = 
\frac{1}{|e| + 1} ( 
\sum_{\scalebox{0.55}{$e \in \{m2a, a2m, m2o, o2m\}$}} H_{oh}(f_{obj}^{e}, \mathcal{Y}) +
H_{oh}(f_{obj}^{mo}, \mathcal{Y})
)
\end{aligned}
\end{equation}
\end{small}

With respect to the visual feature of the composition from image embedder, we calculate the cosine similarities between the visual embedding and the composed pair embeddings of seen composition labels. The cross-entropy loss for the composition is as follows:
\begin{equation}
    \mathcal{L}_{comp} = 
    \frac{1}{|img|} 
    \sum_{img \in \{m, a, o\}} H_{oh}(f_{comp}^{img}, \mathcal{C}_s)
\end{equation}

\subsection{Training and Inference}
During the training phase, the overall loss is as follows:
\begin{equation}
    \mathcal{L} = 
    \gamma_{ortho} \mathcal{L}_{ortho} +
    \gamma_{comp} \mathcal{L}_{comp} + 
    \gamma_{pri}( \mathcal{L}_{attr} + 
    \mathcal{L}_{obj})/2
\end{equation}
where $\gamma_{ortho}$, $\gamma_{comp}$, and $\gamma_{pri}$ are weighting factors to balance the influence of different losses.

For inference, given an image from the test set, the cosine similarities of its visual feature obtained by image embedder and the composed pair embeddings of all candidate compositions are computed. The composition with the highest similarity is predicted by the model. Note that although the disentanglement branch is not used for inference, it still influences the formation of the composition feature space 
through the visual feature extraction module described in Section~\ref{sec:visual-feature-extraction}.
\begin{table}[t]
    \centering
    \LARGE
    \resizebox{1.0\linewidth}{!}{
        \begin{tabular}{lcclcclccclccc}
        \toprule
        \multirow{2}{*}{Dataset}&\multicolumn{2}{c}{\textit{Primitive}}
        &&\multicolumn{2}{c}{\textit{Train}}
        &&\multicolumn{3}{c}{\textit{Validation}}
        &&\multicolumn{3}{c}{\textit{Test}} \\
        \cmidrule{2-3}\cmidrule{5-6}\cmidrule{8-10}\cmidrule{12-14}
        
         & $|\mathcal{A}|$ & $|\mathcal{O}|$ && $|\mathcal{C}_s|$ & $|\mathcal{X}|$ && $|\mathcal{C}_s|$ & $|\mathcal{C}_u|$ & $|\mathcal{X}|$ && $|\mathcal{C}_s|$ & $|\mathcal{C}_u|$ & $|\mathcal{X}|$ \\ \midrule
        MIT-States
        &  115 & 245 && 1262 & 30k && 300 & 300 & 10k && 400 & 400 & 13k \\
        C-GQA
        & 413 & 674 && 5592 & 27k && 1252 & 1040 & 7k && 888 & 923 & 5k \\ 
        VAW-CZSL
        & 440 & 541 && 1252 & 72k && 2121 & 2322 & 10k && 2449 & 2470 & 11k \\
        \bottomrule
        \end{tabular}
    }
    \vspace{-0.5em}
    \caption{Summary statistics of the datasets used in our experiments. }
    \label{tab:split}
\end{table}
\begin{table*}[t]
    \centering
    \resizebox{1.0\linewidth}{!}{
    \begin{tabular}{lllcccclcccclcccc}\toprule
        & \multirow{2}{*}{Method}&& \multicolumn{4}{c}{MIT-States} && \multicolumn{4}{c}{C-GQA} && \multicolumn{4}{c}{VAW-CZSL}\\ \cmidrule{4-7} \cmidrule{9-12} \cmidrule{14-17}
        &  && $AUC$ & $HM$ & $Seen$ & $Unseen$ && $AUC$ & $HM$ & $Seen$ & $Unseen$ && $AUC$ & $HM$ & $Seen$ & $Unseen$ \\ \midrule

    &SymNet~\cite{2020_Li_CVPR_SymNet}
    && 3.2 & 13.7 & 22.7 & 20.1 
    && 1.9 & 10.8 & 20.3 & 11.8 
    && 2.8 & 13.5 & 20.2 & 18.0 \\ 
    
    & CompCos~\cite{2021_Mancini_CVPR_Compcos}
    && 12.3 & 28.2 & 39.0  & 39.5 
    && 5.0  & 17.7 & 32.8 & 19.1 
    && 6.5 & 20.8 & 30.5 & 27.4 \\
    
    & Co-CGE~\cite{2022_Mancini_PAMI_CoCGE}
    && 10.3 & 25.1 & 41.0  & 33.1 
    && 4.2  & 15.2 & 32.9 & 17.0  
    && 6.2 & 19.7 & 31.0  & 26.1 \\ 
    
    & SCEN~\cite{2022_Li_CVPR_SCEN}
    && 9.8 & 24.6 & 35.1 & 36.5 
    && 3.8 & 15.3 & 31.5 & 15.7 
    && 5.7 & 19.2 & 29.9  & 24.5 \\ 
    
    & OADis~\cite{2022_Saini_CVPR_OADis}
    && 13.1 & 29.0  & 42.3 & 27.3 
    && 2.3 & 12.1 & 23.3 & 12.8 
    && 4.1 & 16.2 & 26.0  & 20.7 \\ 
    
    & INV~\cite{2022_Zhang_ECCV_INV}
    && 11.5 & 26.6 & 28.5 & 25.0 
    && 1.4 & 7.9 & 28.6 & 6.8 
    && 2.0  & 11.1 & 21.1 & 11.9 \\ 
    
    & CANet~\cite{2023_Wang_CVPR_CANET}
    && \underline{13.6}  & \underline{29.8} & \textbf{46.4} & 39.9  
    && \underline{5.7} & \underline{18.9} & \underline{34.8} & 20.5 
    && \underline{6.7} & \underline{21.0}  & \underline{31.2} & \underline{27.4} \\ 
    
    & ProCC~\cite{2024_huo_AAAI_procc}
    && 9.5 & 28.1 & 43.1 & 39.1 
    && 3.5 & 15.1 & 32.4 & 15.8 
    && 3.6  & 18.9 & 26.9 & 25.5 \\
    
    \midrule
    
    & CLIP~\cite{2023_Saini_ICLR_CSP}
    && 11.0 & 26.1 & 30.2 & \underline{46.0}  
    && 1.4 & 8.6 & 7.5 & \underline{25.0}  
    && - & - & - & - \\
    
    & CoOp~\cite{2023_Saini_ICLR_CSP}
    && 13.5 & 29.8 & 34.4 & \textbf{47.6} 
    && 4.4 & 17.1 & 20.5 & \textbf{26.8} 
    && - & - & - & - \\ 
    
    \midrule
    
    & \textbf{TRIDENT} (Ours)
    && \textbf{14.2} & \textbf{30.9} & \underline{44.5} & 40.0  
    && \textbf{8.0}  & \textbf{22.6} & \textbf{39.5} & 24.1 
    && \textbf{8.3} & \textbf{23.4} & \textbf{33.3} & \textbf{31.1} \\ 
    
    \bottomrule

    \end{tabular}
    }
    \vspace{-0.5em}
    \caption{
    Comparison with the state-of-the-art results on MIT-States, C-GQA and VAW-CZSL. The four indicators are explained in Metrics. 
    We measure top-1 $AUC$ on MIT-States and C-GQA, and top-3 $AUC$ on VAW-CZSL.
    The best results are displayed in \textbf{boldface}, and the second best results are \underline{underlined}.
    }
    \label{tab:main-results}
\end{table*}

\begin{table}[t]
    \hspace{0.001\textwidth}
    \centering
    \resizebox{1\linewidth}{!}{\begin{tabular}
    {lcccc}
    \toprule
    Method & $AUC$ & $HM$ & $Seen$ & $Unseen$\\
    \midrule
    \textit{w/o patch\_features} & 12.9 & 28.9 & 42.6 & 38.0 \\
    \textit{w/o [CLS]\_feature} & 13.4 & 30.1 & 44.3 & 39.7 \\
    \textit{w/o FAAs} & 13.9 & 30.4 & 44.4 & 39.7  \\
    \textit{w/o condition\_masks} & 14.0 & 30.5 & 44.2 & 39.8 \\
    \textit{w/o word\_expanding} & 14.0  & 30.1 & 44.7 & 39.8 \\
    \textit{w/o attribute\_smoothing} & 13.9 & 30.5 & \textbf{44.9} & 39.5  \\
    \textit{w/o $\mathcal{L}_{attr} + \mathcal{L}_{obj}$} & 13.2 & 30.1 & 43.8 & 38.9 \\
    \textit{w/o $\mathcal{L}_{ortho}$} & 14.1 & 30.7 & 44.6 & 39.7  \\
    \textbf{TRIDENT} & \textbf{14.2} & \textbf{30.9} & 44.5 & \textbf{40.0} \\ %
    \bottomrule
    \end{tabular}}
    \vspace{-0.5em}
    \caption{
    Ablation study results on MIT-States.
    \textit{w/o \{certain\_part\}} denotes this part is ablated.
    }
    \label{tab:ablation-a}
\end{table}

\section{Experiment}
\subsection{Experiment Setup}
\textbf{Datasets.} We evaluate our model on three challenging CZSL datasets: MIT-states~\cite{2015_Isola_CVPR_mit}, C-GQA~\cite{2021_Naeem_CVPR_CGE}, and VAW-CZSL~\cite{2022_Saini_CVPR_OADis}. The common data splits are presented in Table~\ref{tab:split}.

\textbf{Metrics.} Our method is evaluated on seen and unseen compositions separately. Following the common generalized CZSL setting~\cite{2019_Purushwalkam_ICCV_TMN}, a calibration bias trades off between the accuracies of seen and unseen compositions.
We calculate the Area Under the Curve ($AUC$) using seen and unseen accuracies at different biases. The best $Seen$ and $Unseen$ accuracies of the curve are also reported. In addition, we calculate the Harmonic Mean of seen and unseen accuracies at different biases and report the best one ($HM$).

\textbf{Implementation details.} We use the visual encoder of LLaVA v1.5, ViT-Large-Patch14-336px as our frozen visual backbone. \textbf{TRIDENT} and all baseline models are trained with the batch size of 128 for 50 epochs under the PyTorch framework~\cite{2019_pytorch_NIPS}. The number of global features is set to 6, 2, and 4 for the three datasets, respectively, and the number of local features is twice that of the global features. The label smoothing factor is set to 0.09, 0.03, and 0.03 for the three datasets, respectively. The number of auxiliary attributes generated for each composition is set to 3. 
We train \textbf{TRIDENT} by Adam optimizer with the weight decay of 5e-5, learning rates of 1.5e-6 for word embedding and 2e-4 for other modules.
We decay the learning rate by a factor of 0.1 at epoch 30 and 40.
The temperature variable of cosine similarity $\delta$ is set to 0.05.
For weighting coefficients $\gamma_{ortho}$, $\gamma_{comp}$, and $\gamma_{pri}$, we set them to 0.1, 1, 0.25, respectively.

\begin{table}[t]
    \small
    \centering
    \begin{tabular}{lccc}
    \toprule
    Method & $Varient$ & $AUC$ & $HM$ \\
    \midrule
    \multirow{2}{*}{SCEN~\cite{2022_Li_CVPR_SCEN}} 
    & \textit{ft+w2v} & 8.2 & 22.8 \\
    & \textit{LLaVA\textsubscript{lhs}} & \textbf{10.3} & \textbf{25.1} \\
    \midrule
    \multirow{2}{*}{CANet~\cite{2023_Wang_CVPR_CANET}} 
    & \textit{ft+w2v} &12.3 & \textbf{28.4} \\
    & \textit{LLaVA\textsubscript{lhs}} & \textbf{12.5} & 28.3 \\
    \midrule
    \multirow{2}{*}{\textbf{TRIDENT}} 
    & \textit{ft+w2v} &14.0 & 29.9 \\
    & \textit{LLaVA\textsubscript{lhs}} & \textbf{14.2} & \textbf{30.9} \\
    \bottomrule
    \end{tabular}
    \vspace{-0.5em}
    \caption{
    Impact of word embedding on MIT-States. $ft+w2v$ means the sum of Word2Vec and Fasttext. $LLAVA_{lhs}$ represents the last hidden states of LLAVA v1.5.
    }
    \label{tab:ablation-b}
\end{table}

\textbf{Baselines.} We compare our \textbf{TRIDENT} with recent and prominent approaches in CZSL: SymNet~\cite{2020_Li_CVPR_SymNet}, CompCos~\cite{2021_Mancini_CVPR_Compcos}, Co-CGE~\cite{2022_Mancini_PAMI_CoCGE}, SCEN~\cite{2022_Li_CVPR_SCEN}, OADis~\cite{2022_Saini_CVPR_OADis}, INV~\cite{2022_Zhang_ECCV_INV}, CANet~\cite{2023_Wang_CVPR_CANET}, and ProCC~\cite{2024_huo_AAAI_procc}. We replace their visual backbone with ViT-Large-Patch14-336px and retrain all models with the same number of epochs for the sake of fairness. 
In addition, although comparing \textbf{TRIDENT} with CLIP-based methods, which rely on the dual-tower architecture, is very unfair due to inadvertent exposure to unseen compositions, we still choose CLIP~\cite{2021_Radford_ICML_CLIP} and CoOp~\cite{2022_Zhou_CVPR_Coop} as baselines for their strong zero-shot abilities. 

\begin{figure*}[!t]
     \centering
     \begin{subfigure}[t]{0.34\textwidth}
         \centering
         \includegraphics[width=0.99\linewidth]{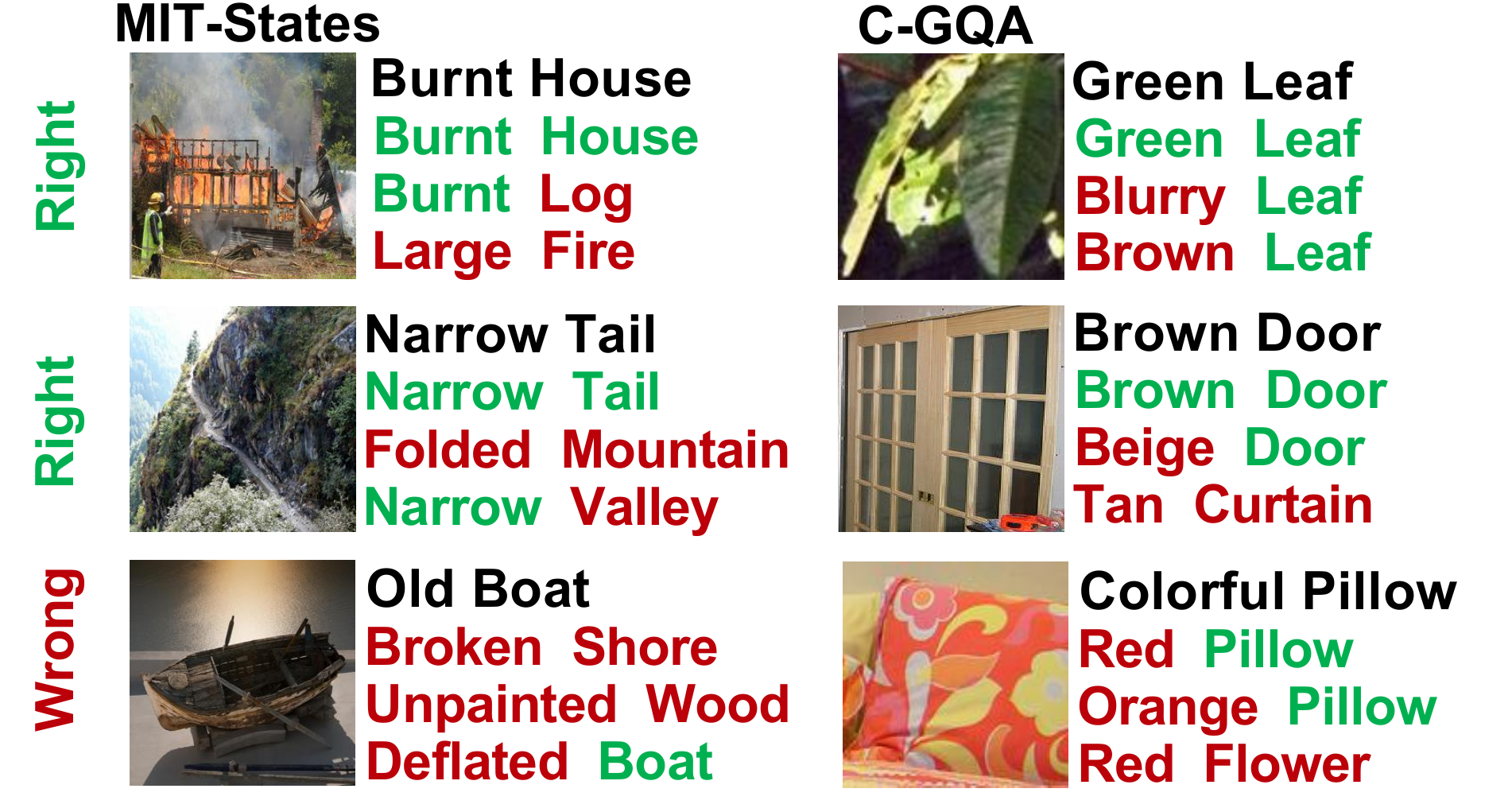}
    \caption{image-to-text retrieval.}
    \label{fig:img2wrd}
     \end{subfigure}
     \hfill
     \begin{subfigure}[t]{0.32\textwidth}
    \centering
    \includegraphics[width=0.9\linewidth]{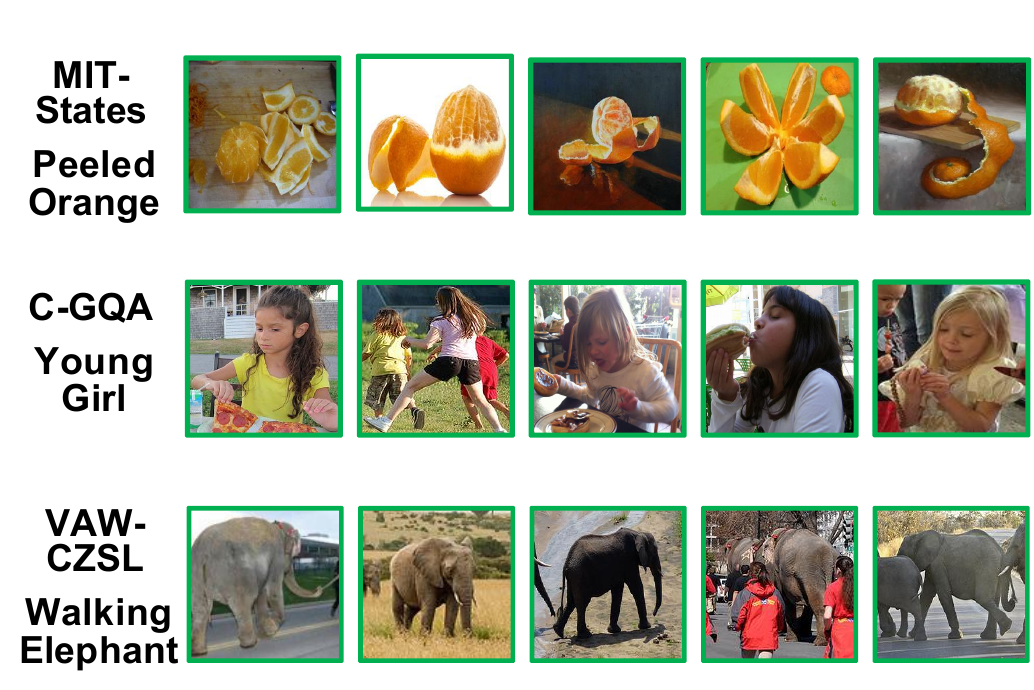}
    \captionsetup{justification=centering}
    \caption{text-to-image retrieval (successful cases).}
    \label{fig:txt2img-r}
     \end{subfigure}
     \hfill
     \begin{subfigure}[t]{0.32\textwidth}
    \centering
    \includegraphics[width=0.9\linewidth]{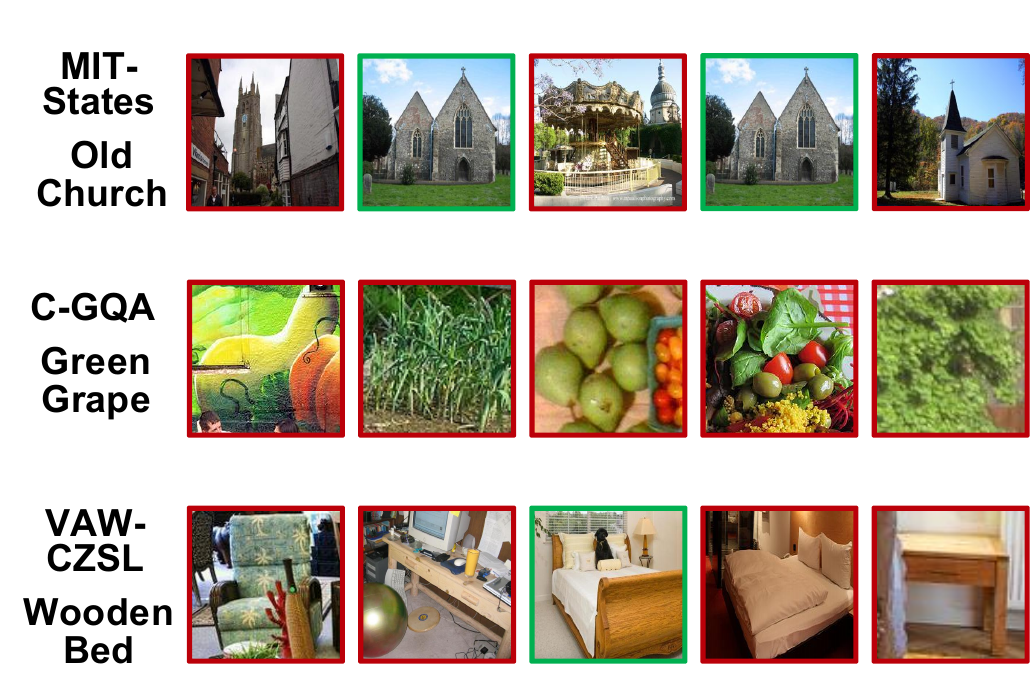}
    \captionsetup{justification=centering}
    \caption{text-to-image retrieval (failure cases).}
    \label{fig:txt2img-w}
     \end{subfigure}
    \caption{Qualitative analysis. (a) Top-5 image-to-text retrieval. The first two rows display successful cases, while the last row presents failure cases. For each image, the top row shows the ground-truth, followed by five rows of top-5 predictions. 
    (b) Successful cases and (c) failure cases of top-5 text-to-image retrieval. 
    In all cases, the successful and failure results are tagged in \textcolor[RGB]{0, 176, 80}{green} and \textcolor[RGB]{187, 0, 9}{red}, respectively.}
    \label{fig:qualitative}
\end{figure*}

\subsection{Results and Discussion}
In this section, we compare \textbf{TRIDENT} with state-of-the-art methods. As shown in Table~\ref{tab:main-results}, our model surpasses other models by a substantial margin in general. \textbf{TRIDENT} boosts $AUC$ from 13.6\%, 5.7\%, and 6.7\% of the previous state-of-the-art method CANet to the new state-of-the-art performance of 14.2\%, 8.0\%, and 8.3\% with 0.6\%, 2.3\%, and 1.6\% improvement on three datasets, respectively. 
In addition, \textbf{TRIDENT} achieves 30.9\%, 22.6\%, 23.4\% on the metrics of $HM$, providing 1.1\%, 3.7\%, and 2.4\% improvement on CANet. 
For MIT-States, our model achieves competitive performance, despite considerable label noise~\cite{2020_NIPS_Atzmon_noise}.
The largest improvement is observed on the $Unseen$ metric, indicating that attribute smoothing helps enhance the generalization ability of the model.
We also observe that \textbf{TRIDENT} performs significantly better than CANet regarding all metrics on two more challenging and low-noise datasets C-GQA and VAW-CZSL, indicating the efficacy of our approach. These improvements arise from the utilization of MLLM embeddings and attribute smoothing, which 
enhance attribute-object disentanglement 
and consequently facilitate the recognition of unseen compositions while maintaining performance on seen compositions.

\subsection{Ablation Study}

We ablate the components of \textbf{TRIDENT} on MIT-States to evaluate their contributions. From the ablation results in Table~\ref{tab:ablation-a}, we gain the following observations. 

1) \textbf{TRIDENT} outperforms the models without using patch and \texttt{[CLS]} features, indicating that both patch and \texttt{[CLS]} features are crucial, with patch features contributing more.

2) Both \textit{w/o FAAs} and \textit{w/o condition\_masks} models perform worse than \textbf{TRIDENT}, which validates the importance of filtering out the background noise and extracting the multi-granularity features, respectively. 

3) \textbf{TRIDENT} surpasses \textit{w/o word\_expanding} model and \textit{w/o attribute\_smoothing} model on the $Unseen$ metric, yet falls short of them on the $Seen$ metric. The difference between our model and the \textit{w/o word\_expanding} model stems from its more thorough disentanglement, which enhances the recognition of unseen compositions while weakening the identification of seen ones. The disparity between our model and the \textit{w/o attribute\_smoothing} model arises from attribute smoothing, which diminishes the overconfidence of the model in seen compositions, facilitating its generalization to unseen ones. However, the improvement of our model over these two models on $AUC$ and $HM$ indicates the effectiveness of the word expanding and label smoothing strategy.

4) \textbf{TRIDENT} outperforms \textit{w/o} $\mathcal{L}_{attr} + \mathcal{L}_{obj}$ model on all metrics, confirming that the attribute-object disentanglement module is highly advantageous for generalization from seen compositions to unseen compositions. 

5) w/o $\mathcal{L}_{ortho}$ model is inferior to \textbf{TRIDENT}, which suggests that the designed orthogonal regularization is helpful to guarantee that different features extract different visual information.

\textbf{Impact of word embeddings.} Our work leverages the last hidden states of LLaVA v1.5 as
word embeddings, while Word2Vec~\cite{2013_mikolov_arxiv_word2vec} and FastText~\cite{2017_fasttext} are the common word embeddings used for MIT-States in previous works. In Table~\ref{tab:ablation-b}, based on three models: SCEN~\cite{2022_Li_CVPR_SCEN}, CANet~\cite{2023_Wang_CVPR_CANET} and \textbf{TRIDENT}, we compare the performance of using the last hidden states of LLaVA v1.5 (${LLaVA}_{lhs}$) and the sum of Word2Vec and FastText (\textit{ft+w2v}), respectively. The results indicate that the last hidden states of MLLM capture more complex multimodal semantic information than ordinary word embeddings.

\subsection{Qualitative Analysis}
In this section, we use \textbf{TRIDENT} to conduct both image-to-text retrieval and text-to-image retrieval experiments on the three datasets.
We first consider image-to-text retrieval, shown in Figure~\ref{fig:img2wrd}. 
For successful cases, such as the image labeled \texttt{burnt house}, we notice that the top four predictions can describe logs burning on a fire in the image. 
In terms of the image labeled \texttt{green leaf}, another successful case, all predicted attributes can describe the \texttt{leaf}, which is due to attribute smoothing learning more attributes for an object. 
For the failure cases, such as the image labeled \texttt{colorful pillow}, the prediction of \texttt{orange pillow} can also describe the image.

We then consider text-to-image retrieval. Successful cases are shown in Figure~\ref{fig:txt2img-r}, while failure cases are shown in Figure~\ref{fig:txt2img-w}. 
We observe that all retrieved images of \texttt{peeled orange} are correct. However, the retrieved images of \texttt{green grapes} are all wrong. This is due to the fact that the training images of \texttt{green grapes} are almost entirely filled with a single grape, which makes it difficult for the model to capture the contour features of a bunch of \texttt{green grapes}.
\section{Conclusion}
In this work, we propose a novel framework termed \textbf{TRIDENT} to address the challenging CZSL task. First, we leverage feature adaptive aggregation modules to mitigate the impact of background, and utilize learnable condition masks to capture
multi-granularity features for attribute-object disentanglement. In addition, we exploit the last hidden states of MLLM to replace ordinary word embeddings, as they can capture complex multimodal semantic information. Moreover, we leverage LLM to generate auxiliary attributes and perform attribute smoothing to diminish overconfidence of the model in seen compositions, which enables it to generalize to unseen compositions more effectively. Extensive experiments conducted on three challenging datasets demonstrate the effectiveness of our method. 

\clearpage
\section*{Acknowledgments}
Our work was supported by the Beijing Natural Science Foundation (No. 4242046).

\bibliographystyle{named}
\bibliography{ijcai25}

\clearpage
\appendix

\section{Auxiliary Attributes Generation by LLM and Post-process}\label{app:1-generation}

As mentioned before, we leverage GPT-3.5~\cite{2023_openai_gpt35} to generate auxiliary attributes for attribute smoothing. The auxiliary attributes are generated based on the composition instead of the object, such as \texttt{ripe apple} and \texttt{sliced apple}, which takes into account the unique contextuality that arises from the combination of attribute and object~\cite{2024_IJCAI_CCZSL}. This ensures that the auxiliary attributes accurately capture the nuances of the specific composition, rather than generating general object-level attributes. Since the generation ability of LLMs is affected by many factors, in this section, we first explore the impact of different prompts. Then we study the influence of the number of auxiliary attributes $t$. In addition, we introduce post-processing of the generated text to eliminate potential noise.

\textbf{Impact of prompt input to LLM.} Users typically interact with LLMs through prompts to request answers to questions, generate text, complete tasks, and more. The model generates text based on the provided prompt, striving to meet the users' requirements~\cite{2024_prompt}. Therefore, the good design of prompt is significant for eliciting knowledge of LLMs, which enables them to better follow our instructions and generate auxiliary attributes with high quality. We first design prompts with different styles, then input them into GPT-3.5 and observe the quality of generated auxiliary attributes. Some prompt examples for the composition \texttt{browned beef} and \texttt{ancient building} are shown in Table~\ref{tab:prompt1} and Table~\ref{tab:prompt2}, respectively.

\begin{table}[!b]

\centering  

\scalebox{0.88}{
\begin{tabular}{l p{6cm} }
\toprule
 
User: & Please give me five adjectives that are most likely to describe the browned beef well.

\\
GPT-3.5: &  1. Flavorful
2. Juicy
3. Savory
4. Tender
5. Rich
\\ \midrule

User: & Now l give you an object with an adjective: browned beef. Please give me five more possible adjectives that can describe the browned beef well. 

\\
GPT-3.5: & 1. Juicy beef
2. Tender beef
3. Flavorful beef
4. Savory beef
5. Succulent beef
\\ \midrule

User: &  Please give me five common adjectives that can describe a browned beef well.

\\
GPT-3.5: & 1. Juicy
2. Brown
3. Savory
4. Tender
5. Succulent
\\ \midrule

User: & Please give me five adjectives that can describe the visual feature of a photo of a/an broned beef well.

\\
GPT-3.5: & 1. Juicy
2. Glistening
3. Crispy
4. Sizzling
5. Mouthwatering
\\
\bottomrule
\end{tabular}
}
\caption{Example prompts with the composition \texttt{browned beef} input to GPT-3.5 and the generated text. }
\label{tab:prompt1} 
\vspace{-1em}
\end{table}

As shown in Table~\ref{tab:prompt1}, the prompt without "the visual feature of ..." may cause the model to produce adjectives that are not specific but generic, such as \texttt{Savory} and \texttt{Rich}. In both Table~\ref{tab:prompt1} and Table~\ref{tab:prompt2}, the prompts starting with "Now I give you...", compared to those starting with "Please give me ...", result in a weaker instruction following ability of the model. Therefore, we choose the prompt: "Please give me $t$ adjectives that can describe the visual feature of a photo of a/an ... well."

Additionally, we incorporate supplementary instructions to constrain the format of the generated text. The complete prompt is as follows: “Please give me $t$ adjectives that can describe the visual feature of a photo of a/an ... well. Please strictly follow the format: give me just $t$ words separated by commas and do not say more.”
\begin{table}[!t]
\centering  

\scalebox{0.88}{
\begin{tabular}{l p{6cm} }
\toprule
 
User: & Please give me five adjectives that are most likely to describe the inflated pool well.

\\
GPT-3.5: &  1. Spacious
2. Durable
3. Sturdy
4. Bright
5. Relaxing
\\ \midrule

User: & Now l give you an object with an adjective: inflated pool. Please give me five more possible adjectives that can describe the inflated pool well. 

\\
GPT-3.5: & 1. Refreshing inflated pool
2. Relaxing inflated pool
3. Blue inflated pool
4. Large inflated pool
5. Fun inflated pool
\\ \midrule

User: &  Please give me five common adjectives that can describe a inflated pool well.

\\
GPT-3.5: & 1. Large
2. Round
3. Inviting
4. Fun
5. Sturdy
\\ \midrule

User: & Please give me five adjectives that can describe the visual feature of a photo of a/an inflated pool well.

\\
GPT-3.5: & 1. Colorful
2. Refreshing
3. Sparkling
4. Bright
5. Relaxing
\\
\bottomrule
\end{tabular}
}
\caption{Example prompts with the composition \texttt{inflated pool} input to GPT-3.5 and the generated text. }
\label{tab:prompt2} 
\end{table}

\textbf{Impact of the number of auxiliary attributes $t$.} In Table~\ref{tab:prompt1}, we observe that the generated attributes describe the compositions to varying degrees, with later items in the sequence being less relevant generally. Therefore, we study the influence of the number of auxiliary attributes $t$.

\begin{table}[t]

\centering  

\scalebox{0.88}{
\begin{tabular}{l p{6cm} }
\toprule

$t$ & the generated text for the composition \texttt{large garden}
\\
\midrule

3 &  1. Lush
2. Vibrant
3. Flourishing

\\ 
\midrule

5 & 1. Lush
2. Expansive
3. Vibrant
4. Serene
5. Verdant

\\
\midrule

10 &  1. Lush
2. Vibrant
3. Expansive
4. Serene
5. Colorful
6. Beautiful
7. Bountiful
8. Captivating
9. Peaceful
10. Tranquil

\\
\bottomrule
\end{tabular}
}
\caption{Impact of $t$ on the generated text with the composition \texttt{large garden}. Note that the input prompt provided to GPT-3.5 is the previously selected one, replacing \textit{t} and the composition.}
\label{tab:t1} 
\end{table}
\begin{table}[htb]

\centering  

\scalebox{0.88}{
\begin{tabular}{l p{6cm} }
\toprule

$t$ & the generated text for the composition \texttt{young girl}
\\
\midrule

3 &  1. Innocent
2. Radiant
3. Youthful

\\ 
\midrule

5 & 1. Youthful
2. Innocent
3. Vibrant
4. Radiant
5. Captivating

\\
\midrule

10 &  1. Radiant
2. Innocent
3. Vibrant
4. Captivating
5. Playful
6. Ethereal
7. Alluring
8. Charming
9. Enchanting
10. Happpy

\\
\bottomrule
\end{tabular}
}

\caption{Impact of $t$ on the generated text with the composition \texttt{young girl}. Note that the input prompt provided to GPT-3.5 is the previously selected one, replacing \textit{t} and the composition.}
\label{tab:t2} 
\end{table}

Table~\ref{tab:t1} and Table~\ref{tab:t2} show the generated text using different $t$ with compositions \texttt{large garden} and \texttt{young girl}. The results demonstrate that the greater the number, the more generic adjectives with irrelevant information are included, for example, \texttt{Captivating} is generated for both compositions. In addition, with $t$ increasing, the noise in the generated text grows due to the uncertainty of the model about the given composition. For example, the \texttt{young girl} may not be happy, yet the model fails to find ten words to describe her, so it has to guess. Therefore, we set $t$ to 3, which minimizes the general adjectives and noise while retaining useful information. 

\textbf{Post-processing of generated text.} GPT-3.5 generates a segment of text, which we need to process into multiple useful words by exploiting regular expressions. However, the auxiliary attributes generated by LLM may contain the attribute of the input composition, for example, generating \texttt{ancient} for \texttt{ancient building}. At this point, we reuse the model to generate $t+1$ adjectives for this composition and select $t$ adjectives that are not the attribute of the input composition.

\section{Auxiliary Attribute Quality Assessment}
In this section, we evaluate the quality of auxiliary attributes in terms of both correctness and diversity on MIT-States. Two approaches are employed: evaluation based on LLM generation and evaluation using CLIP representations. 

Evaluation of the generation quality of LLMs is always performed by querying another LLM multiple times and averaging the results. We assess the quality of auxiliary attributes in terms of correctness and diversity by asking GPT-4o~\cite{2023_GPT4o} with two crafted prompts, respectively: \textit{"Can the word} '[Auxiliary Attribute]' \textit{describe a/an} '[Composition]'\textit{? Please answer 'Yes' or 'No'.}" and "\textit{Does the use of} '[Auxiliary Attribute]'\textit{ to describe a }[Composition]\textit{ provide additional information? Please answer 'Yes' or 'No'.}" For each auxiliary attribute of each combination, we query three times and calculate the average accuracy of the "Yes" responses. As observed in Table~\ref{tab:aux-quality}, the auxiliary attributes are of high quality in terms of both correctness and diversity.
\begin{table}[ht]
    \centering
    \begin{tabular}{lcccc}
    \toprule
        Indicator & Test 1 & Test 2 & Test 3 & Average \\ 
        \midrule
        Correctness & 93.2\% & 92.5\% & 90.8\% & 92.6\% \\ 
        Diversity & 83.5\% & 82.9\% & 83.1\% & 83.2\% \\ 
        \bottomrule
    \end{tabular}
    \caption{Evaluation on the auxiliary attributes on MIT-States.}
    \label{tab:aux-quality}
\end{table}

In addition, we use CLIP and "a photo of" prompt to compare the similarities between the original composition and [auxiliary attribute][object] ([a-a][o]), [random attribute][object] ([r-a][o]), and [auxiliary attribute] "object" ([a-a]"o"), respectively. In Table~\ref{tab:aux-CLIP}, the first similarity greater than the last two respectively indicates auxiliary attributes describe objects correctly, and auxiliary attributes are diverse.

\begin{table}[H]
    \centering

    \begin{tabular}{lccc}
    \toprule
        Experiment & [a-a][o] & [r-a][o] & [a-a]"o" \\ \midrule
        Similarity & 0.8389 & 0.7950 & 0.8197 \\ \bottomrule
    \end{tabular}

    \caption{Similarities of different forms of attribute-objects with the original composition by CLIP.}
    \label{tab:aux-CLIP}
\end{table}

\begin{figure}[H]
  \centering
  \includegraphics[width=0.7\linewidth]{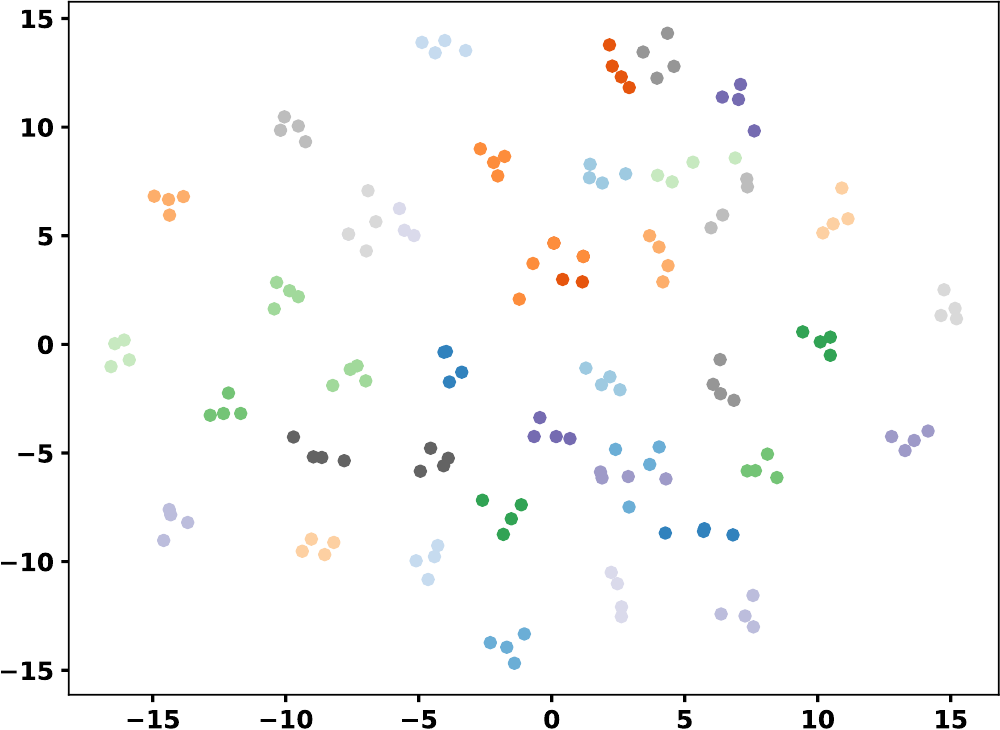}  
  \caption{Visualization of auxiliary attribute-objects ([a-a][o]) and original compositions ([a][o]).}
  \label{fig:aux-tsne-CLIP}
\end{figure}

We also visualize text features of CLIP for $t\ (t=3)$ [a-a][o] and the original [a][o] of randomly selected 40 compositions by t-SNE. In Figure~\ref{fig:aux-tsne-CLIP}, every four features of the composition are in the same cluster (verifying correctness) and the points within the cluster are scattered (verifying diversity).

\section{Obtainment of the Last Hidden States of MLLM}\label{app:2-MLLM}
We input the attribute (object) word into LLaVA-v1.5-7B~\cite{2024_liu_CVPR_llava1.5}, which first tokenizes the word into $o$ tokens. These tokens pass through all Transformer blocks in the MLLM, ultimately generating $o$ embeddings of dimension $d_{m}$ after the last block, named the last hidden states. Subsequently, we apply average pooling to these
$o$ embeddings of dimension $d_{m}$ to obtain a $d_m$-dimensional embedding that represents the attribute. Since the last hidden states are designed to generate the next token rather than for representation, GritLM~\cite{2024_muennighoff_GritLM} leverages instruction to fine-tune the model. Therefore, we fine-tune the last hidden states with a low learning rate during the training phase of \textbf{TRIDENT}.

It is important to note that although LLaVA v1.5 may have seen certain images during training, the model is asked to generate textual descriptions of images in an autoregressive manner during training. The textual descriptions focus on the main content of the image, rather than the "attribute-object" label. 

\section{Dataset Introduction}\label{app:3-data-split}
We evaluate our model on three challenging CZSL benchmark datasets: MIT-States~\cite{2015_Isola_CVPR_mit}, C-GQA~\cite{2021_Naeem_CVPR_CGE}, and VAW-CZSL~\cite{2022_Saini_CVPR_OADis}. 
MIT-States consists of diverse real-world images labeled by early image search engine technology. C-GQA and VAW-CZSL are two more challenging benchmark datasets that consist of broad collections of in-the-wild images. C-GQA has more one-to-one compositions, while objects in VAW-CZSL share more attributes. 
Table~\ref{tab:split} shows detailed data statistics following the common data splits of MIT-States~\cite{2015_Isola_CVPR_mit}, C-GQA~\cite{2021_Naeem_CVPR_CGE} and VAW-CZSL~\cite{2022_Saini_CVPR_OADis}. 
MIT-States contains 53753 images, with 115 attributes and 245 objects. It comprises 1262 seen compositions
and 300/400 (validation/test) unseen compositions.
C-GQA is a natural image dataset which contains 39298 images, with
413 attributes and 764 objects. It includes 5,592 seen compositions and 1,040/923 (validation/test) unseen compositions.
VAW-CZSL is a larger dataset which contains 440 attributes and 541 objects for 238040 images, and it is split into 11175 seen and 2322/2470 unseen compositions for training and validation/testing, respectively.

\section{Impact of Hyperparameters}\label{app:4-hyperpara}
To provide more insight into the effect of visual features and label smoothing, we study the performance of TRIDENT with respect to different numbers of visual features and different label smoothing factors, respectively. Experiments are conducted on two datasets MIT-States and C-GQA. 

\begin{figure}[htb]
    \centering
    \begin{subfigure}{0.23\textwidth}
        \centering
        \includegraphics[width=\linewidth]{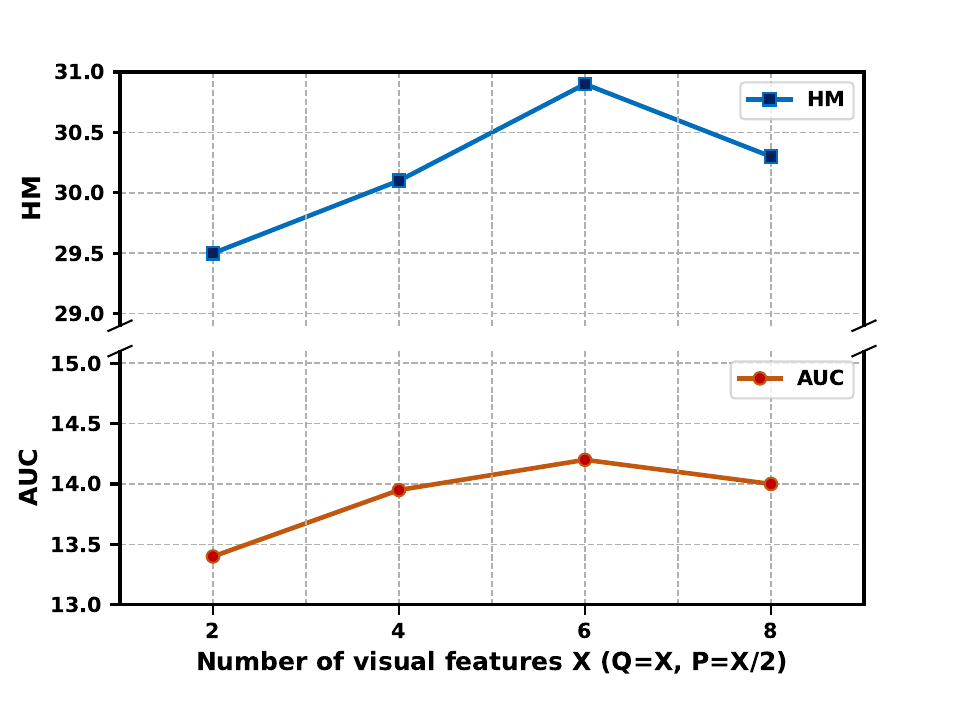}
        \caption{MIT-States}
        \label{fig:p-mit}
    \end{subfigure}
    \begin{subfigure}{0.23\textwidth}
        \centering
        \includegraphics[width=\linewidth]{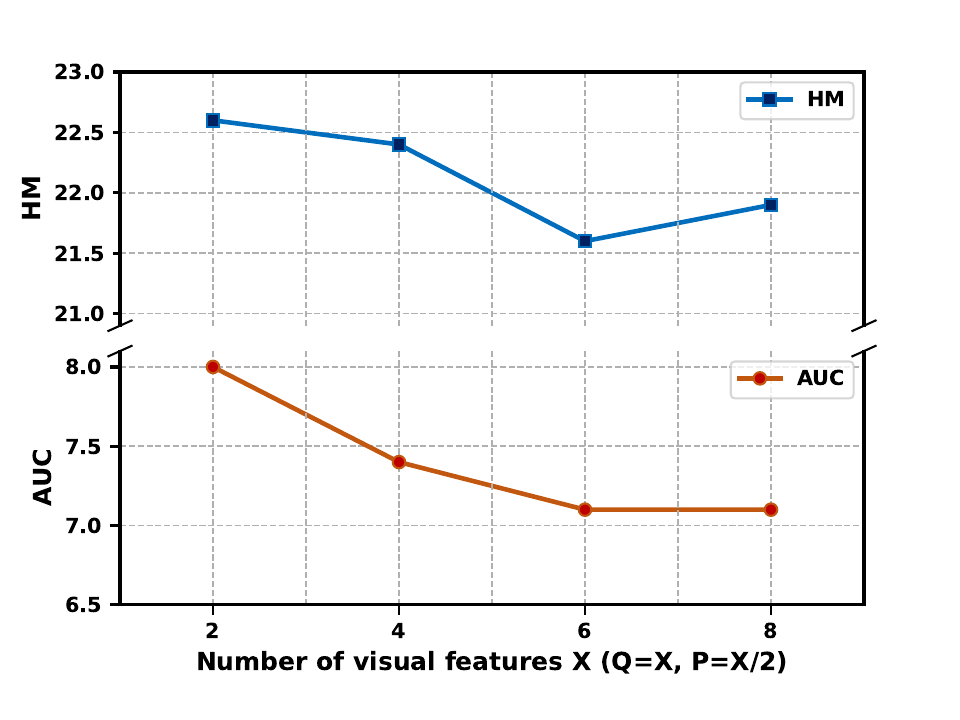}
        \caption{C-GQA}
        \label{fig:p-cgqa}
    \end{subfigure}
    \caption{Impact of the number of the visual features
on (a) MIT-States and (b) C-GQA.}
    \label{fig:p}
\end{figure}

\textbf{Impact of the number of visual features.} In Figure~\ref{fig:p-mit}, the performance of our model on MIT-States generally improves with the increasing number of visual features, but subsequently declines. This trend is reasonable, as a greater number of visual features contain more useful information, thereby enhancing the performance. However, the number of useful features is limited; thus, an excessive number of visual features may introduce redundancy and noise, ultimately hampering the performance of the model. 

However, in Figure~\ref{fig:p-cgqa}, as the number of visual features increases, the performance of the model on C-GQA tends to decline overall. This may be attributed to the model's strong expressive capability in handling composition reasoning. In the low-noise C-GQA dataset, optimal performance can be achieved using only two features. Increasing the number of features, however, results in heightened model complexity without tangible benefits, potentially impairing the performance of the model. In contrast, the MIT-States dataset exhibits significant noise; thus, while the increase in the number of visual features may introduce more noise, it also necessitates a greater amount of useful information, which can effectively mitigate the impact of the noise.

\begin{figure}[htb]
    \centering
    \begin{subfigure}{0.23\textwidth}
        \centering
        \includegraphics[width=\linewidth]{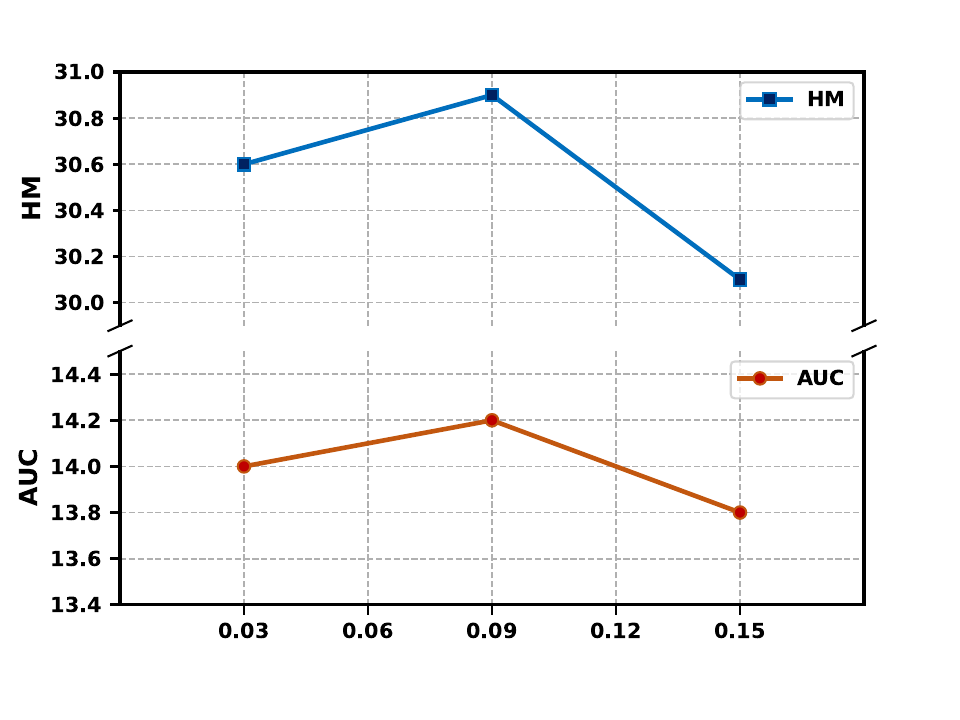}
        \caption{MIT-States}
        \label{fig:alpha-mit}
    \end{subfigure}
    \begin{subfigure}{0.23\textwidth}
        \centering
        \includegraphics[width=\linewidth]{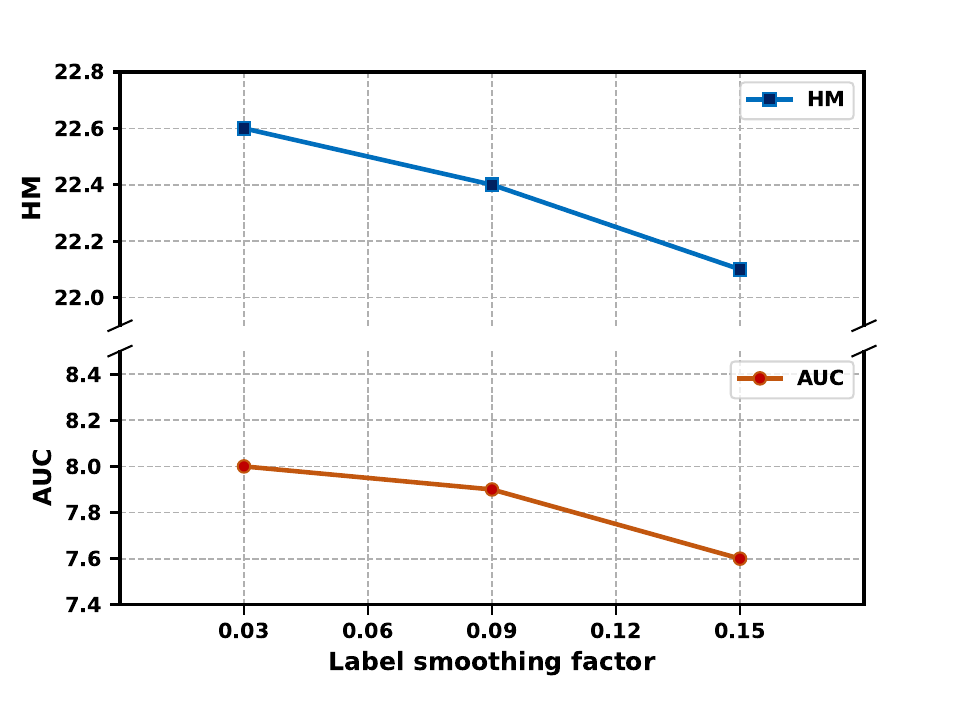}
        \caption{C-GQA}
        \label{fig:alpha-cgqa}
    \end{subfigure}
    \caption{Impact of the label smoothing factor
on (a) MIT-States and (b) C-GQA.}
    \label{fig:alpha}
\end{figure}

\textbf{Impact of the label smoothing factor.} The label smoothing factor $\alpha$ modulates the extent to which the model's confidence in seen compositions is attenuated. Figure~\ref{fig:alpha-mit} shows that as $\alpha$ increases, the model's performance on MIT-States initially improves before subsequently declining. This is because if $\alpha$ is too small, label smoothing fails to enhance generalization, while if alpha is too large, it adversely affects the model's ability to learn the representation of the original labels, resulting in more losses than gains. However, as shown in Figure~\ref{fig:alpha-cgqa}, the model achieves the best performance on C-GQA with a smaller $\alpha$. This may be attributed to the fact that, compared to everyday objects, LLMs are less familiar with in-the-wild objects, leading to relatively lower quality in the generated auxiliary attributes; thus, a smaller smoothing factor can mitigate the impact.

\section{Do Auxiliary Attributes Help or Hurt?}

Ablation study in the full paper proves attribute smoothing is indeed helpful. In addition, MIT-States has 115 attributes, and the set of auxiliary attributes for the dataset has 621 elements. According to statistics, 543 auxiliary attributes (the majority) do not appear in the original dataset. Smoothing with these auxiliary attributes does not mislead the model into mistakenly classifying images into other attributes in the dataset. In addition, changing the smoothing factor $\alpha$ and the number of auxiliary attributes $t$ can control the degree of smoothness to achieve the optimal performance.
\end{document}